\documentclass[letterpaper, 10 pt, journal, twoside]{ieeetran}
\usepackage{amsmath,amsfonts}
\usepackage{algorithmic}
\usepackage{algorithm}
\usepackage{array}
\usepackage[caption=false,font=normalsize,labelfont=sf,textfont=sf]{subfig}
\usepackage{textcomp}
\usepackage{stfloats}
\usepackage{url}
\usepackage{verbatim}
\usepackage{graphicx}
\usepackage{cite}
\usepackage{makecell}
\usepackage{tikz}
\usetikzlibrary{fit, backgrounds, calc, positioning, arrows.meta, shapes.geometric, shadows}
\usepackage{booktabs}
\usepackage{setspace} % For tight line spacing
\newcommand{\btS}[1]{\textcolor{blue!80!black}{\textbf{#1}}} % Blue for [o], {-}, /_/
\newcommand{\btA}[1]{\textcolor{gray!70}{#1}} % Gray for --> and ?
\definecolor{unipdred}{HTML}{9B0014}
\definecolor{unipdgrey}{HTML}{575756}
\definecolor{unipdgreen}{HTML}{167922}

\begin{document}

\title{Real2Sim via Active Perception with Behavior Trees Automatically Generated by VLMs}
%Real2Sim based on Active Perception with automatically VLM-generated Behavior Trees
%Intent-Driven Real2Sim: Autonomous Active Perception via VLM-Generated Behavior Trees

%\author{Anonymous authors
\author{Alessandro Adami$^{*, \dagger}$, Sebastian Zudaire$^{\mathsection}$, Ruggero Carli$^{*}$, Pietro Falco$^{*}$
        % <-this % stops a space
\thanks{Funder: Project co-funded by the European Union – Next Generation Eu - under the National Recovery and Resilience Plan (NRRP), Mission 4 Component 2, Investment 3.3 – Decree no. 630 (24th April 2024)  of Italian Ministry of University and Research; Concession Decree no. 1956 of 05th December 2024 adopted by the Italian Ministry of University and Research, CUP D93D24000270003, within the national PhD Programme in Autonomous Systems (XL cycle)}
% <-this % stops a space
\thanks{\\$*$  University of Padova, Dept. of Information Engineering, Italy.\\
$\mathsection$ ABB Robotics, Västerås, Sweden.\\
$\dagger$ Polytechnic of Bari Dept. of Electrical and Information Engineering, Italy.}
}

% The paper headers
%\markboth{Journal of \LaTeX\ Class Files,~Vol.~14, No.~8, August~2021}%
%{Shell \MakeLowercase{\textit{et al.}}: A Sample Article Using IEEEtran.cls for IEEE Journals}

%\IEEEpubid{0000--0000/00\$00.00~\copyright~2021 IEEE}
% Remember, if you use this you must call \IEEEpubidadjcol in the second
% column for its text to clear the IEEEpubid mark.

\maketitle

\begin{abstract}
Constructing physically accurate simulation environments (Real2Sim) traditionally relies on manual system identification or rigid, exhaustive exploration routines. These task-agnostic pipelines often fail to leverage semantic scene context, leading to redundant physical interactions and inefficient data acquisition. In this paper, we present an autonomous, intent-driven Real2Sim framework that leverages Vision-Language Models (VLMs) for Semantic Task Decomposition. Given a high-level natural language request, an incomplete simulation description, and a visual observation, the framework autonomously identifies the minimal subset of missing physical parameters required for the simulation task. It then generates a reactive Behavior Tree (BT) composed of atomic motion and sensing primitives to selectively acquire these parameters through contact-rich robotic interaction. Extensive real-world experiments on a torque-controlled Franka Emika Panda demonstrate that our approach accurately estimates object mass, surface geometry, and derived parameters such as friction. Quantitative evaluations reveal significant operational efficiency gains compared to exhaustive baseline methods, while ablation studies confirm the robustness of the prompt architecture across different state-of-the-art VLMs. Furthermore, the reactive hierarchy of the BT acts as a deterministic safety filter, successfully mitigating generative VLM hallucinations and preventing unsafe physical anomalies. Ultimately, this work provides a scalable, efficient, and interpretable pipeline for building physics-aware digital twins directly from unstructured human intent.
\end{abstract}


\begin{figure}[]
\centering

\begin{tikzpicture}[
    node distance=0.35cm, % Tightened distances
    % Styles
    header/.style={font=\tiny\bfseries, text=white, rounded corners=1pt, minimum width=3.4cm, minimum height=0.35cm, align=center},
    box/.style={draw=gray!60, fill=white, rounded corners=2pt, minimum width=3.4cm, minimum height=0.6cm, font=\small, align=center, thick, drop shadow={opacity=0.05}},
    fixedbox/.style={draw=gray!80, fill=blue!20, rounded corners=2pt, minimum width=1.6cm, minimum height=0.8cm, font=\small, align=center, thick},
    varbox/.style={draw=blue!60, fill=white, rounded corners=2pt, minimum width=1.6cm, minimum height=0.8cm, font=\small, align=center, thick},
    placeholder/.style={draw=gray!30, fill=white, rounded corners=1pt, font=\fontsize{6}{7}\selectfont, align=center},
    arrow/.style={-{Stealth[scale=0.8]}, line width=1.3pt, draw=gray!80},
    % Group Styles
    group_bg/.style={draw=gray!15, fill=#1, rounded corners=3pt, inner sep=3pt}
]

% --- STAGE 1: INPUTS ---
\node[header, fill=blue!70] (h_in) {\fontsize{9}{10}\selectfont1. INPUT LAYER};

% Side-by-side Fixed vs Variable
\node[varbox, below=0.3cm of h_in] (user) {USER INPUTS\\$\mathcal{R}/\mathcal{D}$ \\\textit{(Agnostic User)}};
\node[fixedbox, left=0.8cm of user] (sys) {FIXED SYSTEM\\PROMPT $\mathcal{S}$\\\textit{(Expert Config)}};
\node[varbox, right=0.8cm of user, yshift=0.2cm] (thumb_in) {IMAGE INPUT\\$\mathcal{I}$ \\ \\\includegraphics[width=2.0cm]{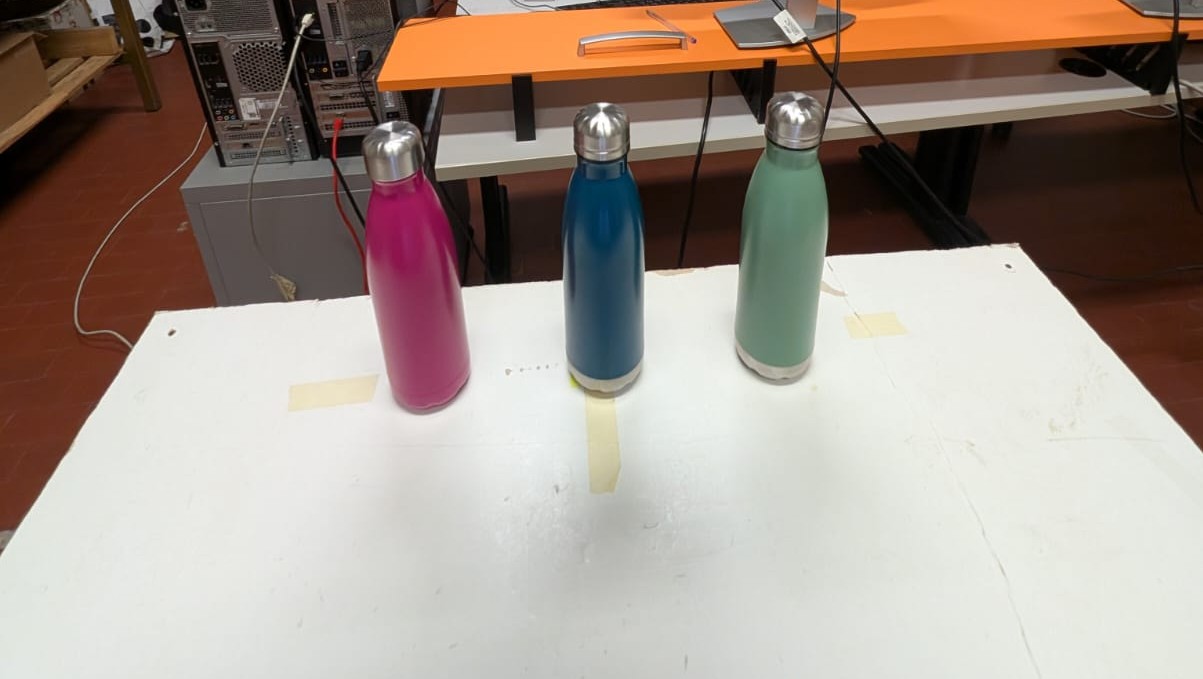}};

% Mixing Junction
\node[circle, draw=gray!60, fill=gray!10, font=\small, inner sep=1pt, below=2.5cm of h_in] (mix) {+};

\draw[arrow] (sys.south) -- ++(0,-0.61) -| (mix);
\draw[arrow] (user.south) -- ++(0,-0.15) -| (mix);
\draw[arrow] (thumb_in.south) -- ++(0,-0.28) -| (mix);

\begin{scope}[on background layer]
    \node[group_bg=blue!5, fit=(h_in) (sys) (thumb_in) (mix)] (g1) {};
\end{scope}

% --- STAGE 2: FRAMEWORK ---
\node[header, fill=unipdred!70, below=3.2cm of h_in] (h_frame) {\fontsize{9}{10}\selectfont2. AUTONOMOUS FRAMEWORK};
\node[box, below=0.2cm of h_frame] (vlm) {VLM Reasoning: Semantic Scene \\Analysis\& Minimal Parameter \\Discovery ($\Phi$)};
\node[box, below=0.3cm of vlm] (bt) {BT Generation \& Reactive Planning};

% Thumbnails tucked in to save width
\node[placeholder, right=-0.05cm of bt, yshift=0.2cm, inner sep=2pt, draw=unipdred!40, fill=white, font=\fontsize{6}{7}\selectfont, align=left] (thumb_bt) {
\textbf{BT Generation:}\\ \\
    \btS{\{--\}} Sequence\\
    \hspace{5pt}\btA{$\rightarrow$} Action 1\\
    \hspace{5pt}\btA{$\rightarrow$} Action 2\\
    \hspace{5pt}\btA{$\rightarrow$} Measure 1\\
    \hspace{5pt}\btA{$\rightarrow$} Action 3\\
    \hspace{5pt}\btS{/\_/} Parallel\\
    \hspace{10pt}\btA{$\rightarrow$} Action 4\\
    \hspace{10pt}\btA{$\rightarrow$} Measure 2\\

};
\node[box, below=0.3cm of bt] (exec) {Robot Execution: Atomic Primitives $\mathcal{A}$};
\node[placeholder, left=-0.05cm of exec, xshift=-0.0cm, yshift=0.6cm] (thumb_rob) {\includegraphics[width=1.5cm]{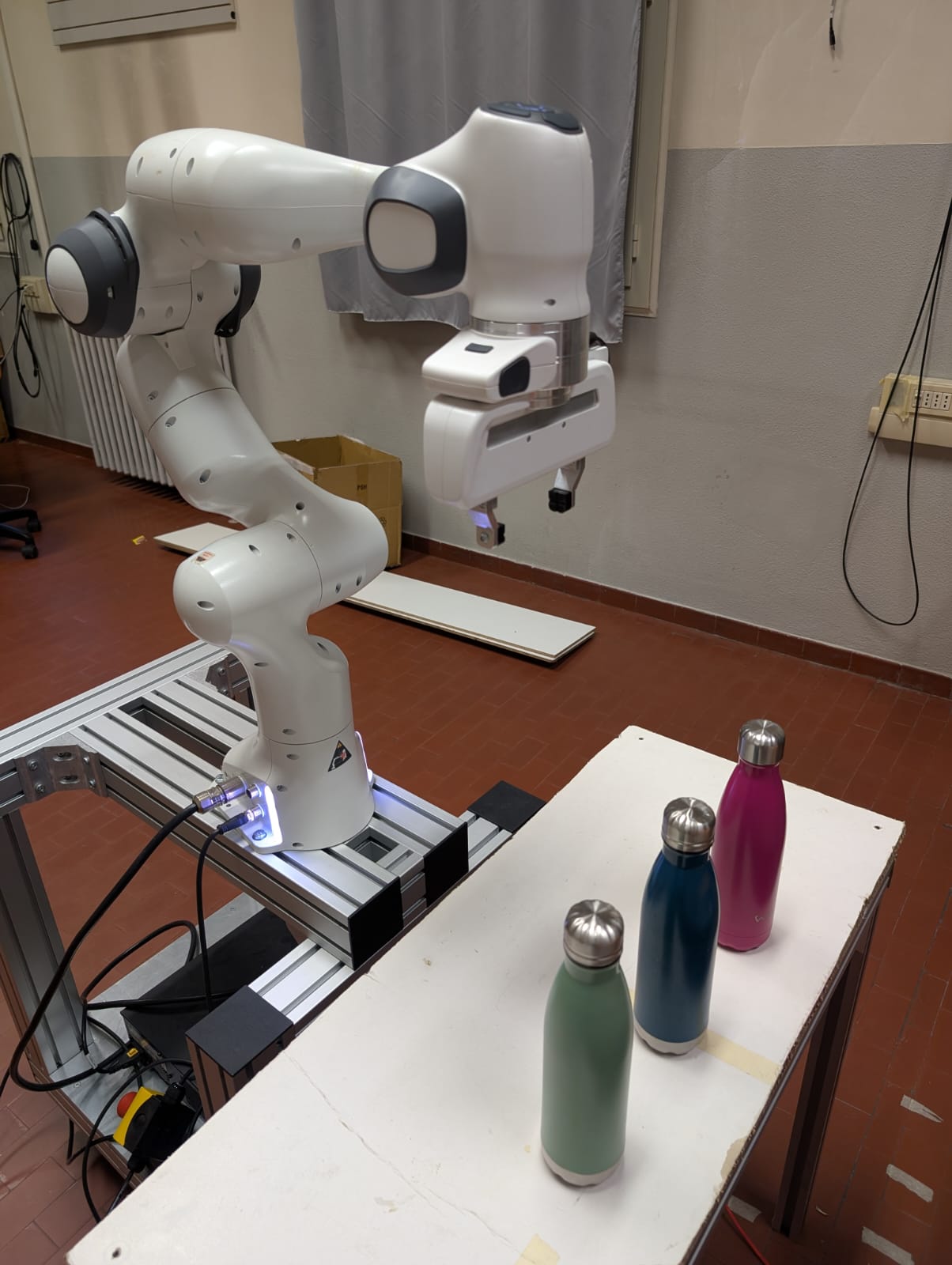}};

\draw[arrow] (vlm) -- (bt);
\draw[arrow] (bt) -- (exec);

\begin{scope}[on background layer]
    \node[group_bg=unipdred!10, fit=(h_frame) (vlm) (exec) ] (g2) {};
\end{scope}

% --- STAGE 3: OUTPUT ---
\node[header, fill=unipdgreen!60, below=3.8cm of h_frame] (h_out) {\fontsize{9}{10}\selectfont3. OUTPUTS};

\node[box, below=0.2cm of h_out, xshift=-2.5cm, yshift=-0.5cm] (est) {Parameter Estimation $\Phi$\\(Mass, Height, Friction...)};
\node[placeholder, right=0.2cm of est] (sim) {\includegraphics[width=2.0cm]{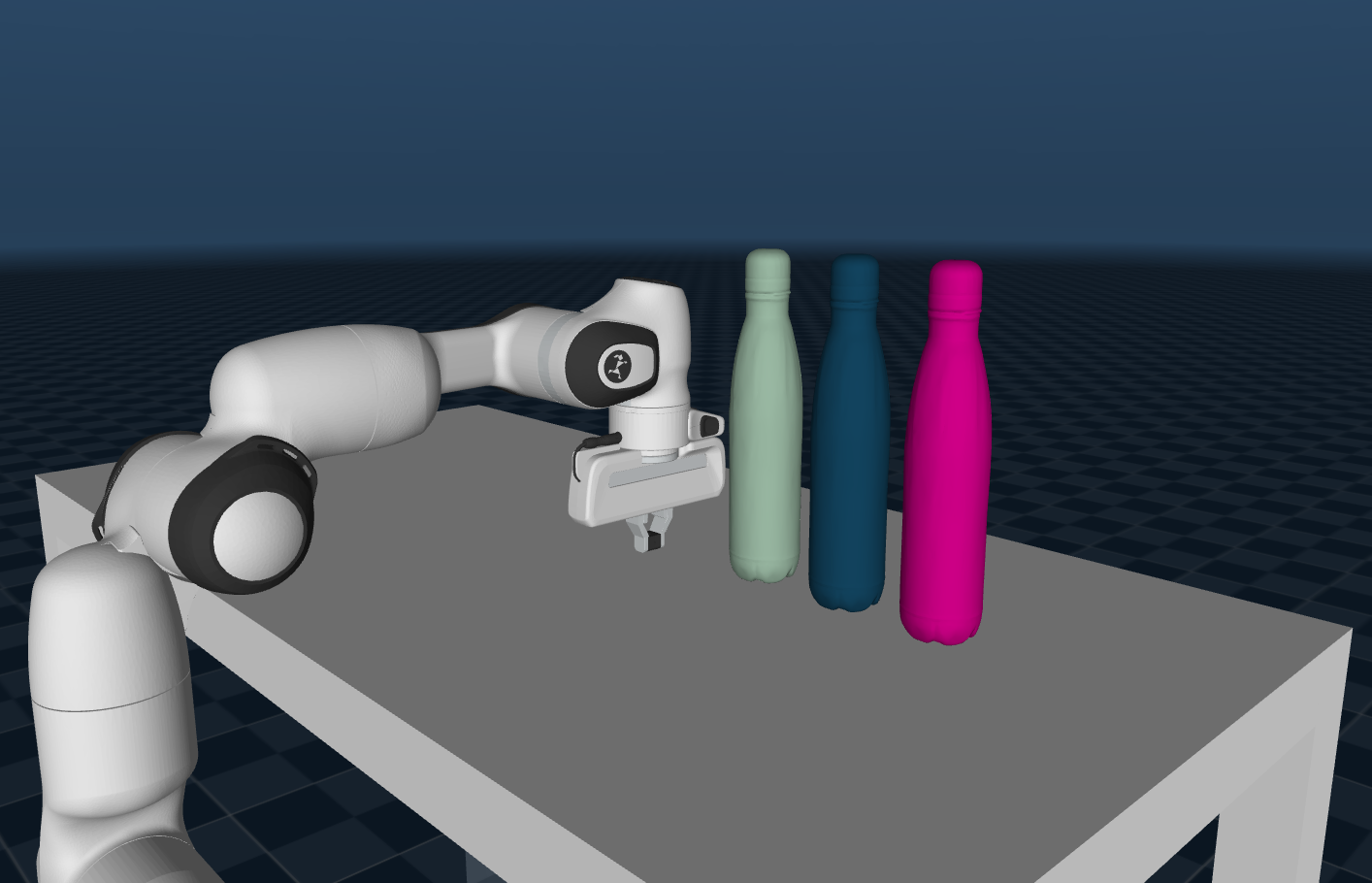}\\ \small Physics-Aware MuJoCo Model};

\begin{scope}[on background layer]
    \node[group_bg=unipdgreen!10, fit=(h_out) (est) (sim)] (g3) {};
\end{scope}

% Main Flow Connectors
\draw[arrow] (mix) -- (h_frame.north);
\draw[arrow] (exec.south) -- (h_out.north);

\end{tikzpicture}

\caption{Compact Real2Sim adaptive framework. The user specifies the desired objective, while the system prompt (fixed and not accessible to the final user) explains how to deal with available atomic functions and the desired output. Then the VLM returns the Behavior Tree for the acquisition of the missing parameters. }
\label{fig: framework}
\end{figure}

\section{Introduction}
\IEEEPARstart{C}{reating} accurate, physically grounded simulations is a fundamental challenge in robotics. Digital twins are essential for safely testing control policies and training reinforcement learning agents prior to real-world deployment~\cite{ROSEN2015567}. However, high-fidelity simulators (e.g., MuJoCo~\cite{todorov2012mujoco}) require precise physical parameters such as mass, center of mass, surface friction, and contact geometry. Because these dynamic properties are often unknown \emph{a priori} and cannot be reliably inferred from purely visual observations~\cite{estobjinertia}, robots must actively interact with their environment to acquire them.

Existing real-to-simulation (Real2Sim) pipelines automate this parameter estimation, but they typically rely on rigid, pre-programmed interaction routines. A significant limitation of current state-of-the-art approaches~\cite{pfaff2025scalablereal2simphysicsawareasset, heiden2019real2sim} is their task-agnostic nature: they exhaustively estimate all identifiable physical parameters regardless of the downstream simulation objective. This exhaustive methodology leads to redundant, time-consuming robotic interactions and fails to leverage semantic scene context. More critically, these systems do not reason over \emph{user intent}, creating a disconnect between high-level simulation goals and low-level robotic exploration.

Simultaneously, vision-language models (VLMs) have demonstrated exceptional capabilities in grounding natural-language instructions into robotic actions~\cite{kim24openvla}. Yet, translating VLM outputs directly into unstructured action sequences lacks the deterministic logic required for contact-rich, physical interactions. While Behavior Trees (BTs) offer a modular and interpretable formalism to ensure reactive execution~\cite{Colledanchise_2018}, the integration of VLMs to autonomously sequence \emph{low-level physical primitives} remains largely unexplored. Current VLM-to-robot frameworks typically rely on opaque, high-level skills (e.g., "pick-and-place") that lack the granular, force-aware sensing necessary for precise system identification.

To bridge this gap, we propose an autonomous, intent-driven Real2Sim framework (Fig.~\ref{fig: framework}) driven by \textbf{Semantic Task Decomposition}. Given a high-level user request, an incomplete simulation description, and an RGB observation, a VLM identifies the minimal subset of missing physical parameters. It then autonomously synthesizes a reactive BT composed solely of atomic robotic primitives. Rather than generating a rigid script, the system constructs a logical hierarchy that deploys sensing actions only when necessary to fill specific "information gaps." This BT is subsequently executed on a torque-controlled Franka Emika Panda, where its reactive structure serves as a deterministic safety layer against generative VLM hallucinations.

The core contributions of this work are threefold:
\begin{itemize}
    \item \textbf{Intent-Driven Active Perception Pipeline:} We introduce a Real2Sim framework that conditions parameter acquisition on user intent and semantic scene understanding. This selective estimation eliminates redundant interactions, vastly improving operational efficiency over exhaustive baseline methods.
    \item \textbf{Autonomous BT Synthesis from Atomic Primitives:} We present a methodology for generating executable, reactive BTs directly from a vocabulary of low-level motion and sensing actions (e.g., Cartesian probing, force logging). By avoiding "black-box" high-level skills, our approach enables the fine-grained, contact-aware interaction required for physical parameter estimation.
    \item \textbf{Empirical Validation of LLM Safety and Robustness:} We validate the framework on a physical manipulator across complex scenarios involving occlusions and derived parameters (e.g., friction). Furthermore, we provide a quantitative ablation study demonstrating how structured prompt engineering, combined with BT logic, ensures structural reliability and safe execution.
\end{itemize}

By explicitly coupling high-level multi-modal reasoning with structured robotic interaction, this work advances toward scalable, autonomous, and interpretable Real2Sim pipelines for unstructured real-world environments.

\textbf{To support reproducibility and future research in intent-driven simulation, the complete software framework and prompt will be made publicly available.}

\section{Related Works}
\subsection{Real2Sim and Digital Twin Construction}\label{sec: r2s related works}
High-fidelity digital twins support simulation-based planning under realistic dynamics~\cite{tao2019}. While recent automated pipelines advance the visual and spatial generation of 3D environments~\cite{ling2025scenethesislanguagevisionagentic, wang2025embodiedgengenerative3dworld, liu2025agentic3dscenegeneration}, they largely ignore underlying physical parameters like mass or friction. Traditional Real2Sim approaches estimate these properties via robotic interaction~\cite{pfaff2025scalablereal2simphysicsawareasset, heiden2019real2sim}, but they typically rely on rigid, fixed exploration routines. They lack adaptation to user goals or scene context and rarely leverage semantic reasoning to guide interaction, limiting their utility in unstructured scenarios.

\subsection{Behavior Trees and VLM-Based Robotic Generation}
Vision-Language Models (VLMs) and Vision-Language-Action (VLA) models, such as OpenVLA~\cite{kim24openvla}, PaLM-E~\cite{driess2023palme}, and RT-2~\cite{brohan2023rt2visionlanguageactionmodelstransfer}, ground instructions in visual observations to produce robot-executable actions. Behavior Trees (BTs) complement this by providing a modular, reactive, and interpretable formalism for task execution~\cite{Colledanchise_2018, IOVINO2022104096}. 
Recent work utilizes LLMs and VLMs to generate BTs from natural language~\cite{wake2025vlm-driven, lykov2023llmbrainaidrivenfastgeneration} or integrates them with reactive planners for failure recovery~\cite{ahmad2025unifiedframeworkrealtimefailure}. However, these approaches generally focus on task execution using predefined, high-level skills with fixed semantics. Unlike prior work assuming actions like \emph{PickUp}~\cite{wake2025vlm-driven}, our framework plans directly over atomic motion and sensing primitives, enabling autonomous, intent-driven parameter acquisition for Real2Sim.

\subsection{Active System Identification and Exploration}\label{Sec: Sys id}
Prior research optimizes physical interactions using Bayesian inference to identify informative data points~\cite{9341401, Kim_2025} or Reinforcement Learning policies to actively reveal occluded inertial parameters~\cite{margolis2023learningphysicalpropertiesactive, memmel2024asidactiveexplorationidentification}. While these methods achieve high mathematical precision, they remain largely task-agnostic, optimizing purely for numeric information gain rather than downstream semantic utility. In contrast, our framework introduces \textbf{Intent-Driven Active Perception}. By grounding natural language in visual observations, our VLM-based Semantic Task Decomposition reasons about task requirements to acquire only the minimal, task-relevant subset of parameters. This yields an efficient, interpretable exploration strategy that adapts zero-shot to varying user needs without requiring computationally expensive policy retraining.
%Prior research has focused on optimizing physical interactions through mathematical or learned heuristics. For instance, specific approaches such as Online BayesSim~\cite{9341401} and ASBI~\cite{Kim_2025} utilize Bayesian inference and indicator metrics to identify the most informative data points for reducing the sim-to-real gap. 
%Other approaches leverage learned motor policies to disambiguate physical properties that are occluded from vision~\cite{margolis2023learningphysicalpropertiesactive}. Specifically, Active Exploration for System Identification (ASID)~\cite{memmel2024asidactiveexplorationidentification} employs Reinforcement Learning to develop exploration policies that move objects to reveal their inertial parameters. While these methods achieve high mathematical precision, they are typically task-agnostic and operate on purely numeric representations of the scene, optimizing for information gain rather than downstream semantic utility. 
%In contrast, our framework introduces Intent-Driven Active Perception. Unlike purely numeric or mathematically exhaustive approaches, our VLM-based method performs Semantic Task Decomposition. By grounding natural language in visual observations, the system reasons about object affordances and task requirements to select only the minimal, task-relevant subset of physical parameters to estimate. This allows for a more efficient and interpretable exploration strategy that adapts to varying user needs without the computationally expensive requirement of retraining task-specific policies.
\\\\
In summary, while prior work has advanced physical parameter estimation, visual 3D generation, VLM-based planning, and active system identification in isolation, no existing framework integrates semantic user intent with low-level, atomic robotic interactions. Our framework bridges this gap, creating the first intent-driven pipeline capable of selectively estimating physical parameters for simulation construction.

\section{Problem Formulation and Solution}\label{sec: Methodology}
Our framework constructs a simulation-ready model by autonomously combining human intent, multi-modal scene understanding, behavior-tree-guided robotic exploration, and physics-based parameter estimation. The overall pipeline transforms a high-level user request into a combination of targeted interaction behaviors that acquire the physical parameters required for simulation, estimating missing parameters only. Notably, the same system prompt is reused unchanged across all experimental scenarios in Sec.~\ref{sec:experiments}, demonstrating that task adaptation emerges from VLM multi-modal reasoning rather than from manual task-specific tuning.

\subsection{Problem Formulation}\label{Sec: problem}
We consider a real-world scene containing objects with partially unknown physical properties. A human operator provides an incomplete simulation description $\mathcal{D}$ (e.g., a MuJoCo~\cite{todorov2012mujoco} \texttt{.xml} file) and/or a high-level natural language request $\mathcal{R}$, which specifies the simulation goal (e.g., \textit{"estimate the weight of the bottle"}) rather than a specific measurement procedure. The user is assumed to be agnostic to robotics and Behavior Trees (BTs).

Given the context tuple $(\mathcal{D}, \mathcal{R}, \mathcal{I})$, where $\mathcal{I}$ is an RGB observation of the workspace, the objective is to construct a complete, physics-aware simulation model $\mathcal{M}$:
\begin{equation}
    \mathcal{M} = \langle \mathcal{G}, \mathcal{P} \rangle
\end{equation}
where $\mathcal{G}$ denotes the geometric configurations (Sec.~\ref{sec: r2s related works}) and $\mathcal{P}$ denotes the full set of physical dynamics parameters (mass, friction, height, etc.). 

Unlike traditional Real2Sim approaches that use fixed exploration routines to estimate the entirety of $\mathcal{P}$, our system performs \textbf{Semantic Task Decomposition}. It maps the high-level intent to identify the minimal, task-relevant subset of missing parameters $\Phi \subseteq \mathcal{P}$, and generates a reactive exploration policy to acquire only $\Phi$. Let $\mathcal{P}_{\text{known}} \subseteq \mathcal{P}$ be the subset of physical parameters already provided in the simulation description $\mathcal{D}$. The complete physical parameter set required for the model $\mathcal{M}$ is achieved through the set union:
\begin{equation}
    \mathcal{P} = \mathcal{P}_{\text{known}} \cup \Phi, \;\;\;\text{with}\;\;\; \mathcal{P}_{\text{known}} \cap \Phi = \emptyset 
\end{equation}
where $\Phi$ is autonomously acquired by the VLM-generated Behavior Tree. 

While acquiring precise 6D poses in unstructured environments remains an active area of computer vision research, this paper focuses specifically on the downstream challenge of semantic reasoning and physical parameter extraction. Therefore, to isolate and validate our intent-driven VLM-BT architecture, we treat initial object poses $p_{i}$ as a modular upstream input provided as environmental metadata. In a fully deployed system, this pipeline would naturally ingest the output of a standard 6D pose estimator.

\subsection{VLM-Based Semantic Task Decomposition and Plan Generation}\label{Sec: scene undertsanding}
The core of the framework is a mapping function $f$ that translates the current state and intent into an executable Behavior Tree:
\begin{equation}
    f: (\mathcal{S}, \mathcal{D}, \mathcal{R}, \mathcal{I}) \rightarrow \mathcal{BT}(\mathcal{A}, \mathcal{C})
\end{equation}
where $\mathcal{S}$ is the fixed and engineered system prompt. 
A fundamental design choice is planning over a general-purpose vocabulary of atomic primitives $\mathcal{A}$ rather than task-specific skills (e.g., \texttt{PickAndWeigh}). While measurement primitives like \texttt{Force2Mass} encapsulate engineered data-processing heuristics, they represent a one-time, hardware-level expert configuration. Once defined, the VLM dynamically recombines these task-agnostic building blocks into novel exploration strategies across any unstructured environment, bypassing the rigid template-matching of traditional high-level skill execution. This provides three critical advantages for autonomous Real2Sim. Similar to how a finite set of letters forms an infinite number of words, the atomic actions in $\mathcal{A}$ can be composed into novel exploration strategies that the system designer may not have explicitly pre-defined.
%A fundamental design choice in this work is planning over a \textbf{low-level atomic action set $\mathcal{A}$} in place of high-level, task-specific skills (e.g., \textit{PickAndWeigh}). While high-level skills simplify the planning, they restrict the robot to a template-matching regime, executing pre-programmed, rigid routines. By contrast, our framework utilizes the VLM as a heuristic planner over a \textbf{Robotic Vocabulary} of elementary motion and sensing primitives. This provides three critical advantages for autonomous Real2Sim. Similar to how a finite set of letters forms an infinite number of words, the atomic actions in $\mathcal{A}$ can be composed into novel exploration strategies that the system designer may not have explicitly pre-defined.

To formalize the concept of the Behavior Tree as a \textbf{Deterministic Safety Filter}, we model the generated $\mathcal{BT}$ not merely as an execution sequence, but as a discrete-time verification policy. Let $s_t$ denote the real-time physical state of the robot (e.g., joint states, torques, Cartesian errors) at control tick $t$. Every atomic action $a \in \mathcal{A}$ inherently evaluates a hardware-level safety predicate $p_{\text{safe}}(s_t)$ (e.g., verifying that a VLM-generated target pose is kinematically feasible). 
By structurally routing the output through the composite execution logic $c \in \mathcal{C}$, the framework guarantees that generative hallucinations cannot bypass physical limits. An anomalous VLM command deterministically returns \texttt{FAILURE} at the leaf node, triggering a safe control abort and creating a rigorous physical boundary between semantic reasoning and hardware execution.

%Similar to how a finite set of letters forms an infinite number of words, the atomic actions in $\mathcal{A}$ can be composed into novel exploration strategies that the system designer may not have explicitly pre-defined. Mapping directly to atomic primitives allows the Behavior Tree to inject \textbf{Condition Nodes} $\mathcal{C}$ at any stage of physical interaction. This ensures the robot can react to contact-rich events, such as a grasp slip or unexpected collision, with a temporal and logical granularity impossible in monolithic, black-box skills. By decomposing the task into primitives, the VLM can selectively omit unnecessary sub-actions based on user intent. It actively discards useless conversion actions from the vocabulary (e.g., ignoring \texttt{Force2Velocity} when evaluating statics) and composes the remaining base primitives into more complex, targeted measurement chains. If $\mathcal{D}$ indicates that an object's geometry is known, the model bypasses surface-probing entirely.

\begin{table}[]
\centering
\footnotesize
\caption{Atomic Action Set $\mathcal{A}$ (Robotic Vocabulary). All the actions will be denoted with the symbol \small$\rightarrow$.}
\label{tab:actions}
\begin{tabular}{@{}lll@{}}
\toprule
\textbf{Action} & \textbf{Input/Output} & \textbf{Utility} \\ \midrule
MovePose & \textbf{I:} Target Pose & \makecell{Cartesian motion to an\\ equilibrium pose} \\ \midrule
MoveJoints & \textbf{I:} Joint Target & Configuration-space motion \\ \midrule
Open/CloseGripper & None & \makecell{Control of the parallel-jaw\\ gripper} \\ \midrule
MoveUntilContact & \makecell{\textbf{I:} direction of \\ movement}& \makecell{Compliant moving along one\\ axis chosen by the VLM} \\ \midrule
MeasureForces & \textbf{O:} gripper\_forces & \makecell{Acquisition of external wrench \\$\boldsymbol{w}_{ext}$ buffered in gripper\_forces} \\ \midrule
MeasureGripperPose & \textbf{O:} gripper\_pose & \makecell{Acquisition of gripper tips\\ pose buffered in gripper\_pose} \\ \midrule\midrule
Force2Mass & \textbf{I:} gripper\_forces & \makecell{Processes the arg and \\ retrieves the mass of the object} \\ \midrule
Force2Velocity & \textbf{I:} gripper\_forces & \makecell{Processes the arg and retrieves\\ the velocity of the gripper} \\ \midrule
Force2Momentum & \textbf{I:} gripper\_forces & \makecell{Processes the arg and retrieves\\ the momentum of the gripper} \\ \midrule
Pose2Height & \textbf{I:} gripper\_pose & \makecell{Processes the arg and retrieves \\ the $z$ pose of the gripper} \\ \midrule
Pose2Plan & \textbf{I:} gripper\_pose & \makecell{Processes the arg and retrieves\\ the $x-y$ pose of the gripper} \\\bottomrule
\end{tabular}
\end{table}

The VLM processes the input through three logical stages:
\begin{enumerate}
    \item \textbf{Intent Grounding and Parameter Discovery:} The VLM identifies scene objects $\mathcal{O}$ referenced in $\mathcal{R}$ or $\mathcal{D}$ and infers the minimal set of required physical parameters $\Phi$ needed to complete $\mathcal{M}$. By comparing the request with available data, the VLM identifies the exact information gap.
    \item \textbf{Action Selection:} The VLM maps $\Phi$ to a sequence of atomic primitives from the predefined library $\mathcal{A}$ (Tab.~\ref{tab:actions}). These actions represent a general-purpose robotic vocabulary for sensing (e.g., measuring forces) and interaction (e.g., moving, activating the gripper). Crucially, sensing primitives within $\mathcal{A}$ act as implicit Condition Nodes. For example, \texttt{CloseGripper} returns \texttt{SUCCESS} only if a physical resistance is met, allowing a parent control node to gracefully handle a failed action without halting the entire control loop.
    \item \textbf{Structured BT Synthesis:} To ensure the plan is reactive and robust, the VLM composes the selected actions using composite nodes $\mathcal{C}$ (Tab.~\ref{tab:composites}). The VLM synthesizes a structured output:
    \begin{equation}
        \mathcal{BT} = [\, \textsf{node}_1, \textsf{node}_2, \dots \, ], \quad \textsf{node}_i = \big(\textsf{args},\textsf{type}\big)
    \end{equation}
    where $\textsf{type} \in \{\mathcal{A}, \mathcal{C}\}$. This ensures the generated behavior is interpretable, reproducible, and strictly compatible with standard BT parsers.
\end{enumerate}

The choice of BTs as the execution formalism is motivated by their reactivity. At each control cycle $t$, the tree is ticked, allowing for an immediate response to sensor feedback. This is particularly critical for contact-rich parameter acquisition; if an action fails, a control node can safely trigger a recovery branch or halt execution, acting as a deterministic safety filter against VLM hallucinations. The model is also instructed to output an explanation text $\mathcal{E}$ detailing the reasoning behind its chosen strategy.

\subsection{Prompt Composition}\label{sec: system prompt}
To minimize human intervention during execution, we strictly separate task-agnostic expert configuration from task-specific user input.

\subsubsection{\textbf{System Prompt $\mathcal{S}$ (The Grammar)}}
The system prompt is a fixed, pre-engineered instruction set defining the robot's capabilities and safety limits. Designed once by an expert, it remains unchanged across all task executions. We decompose it into six key functional modules to provide the model with rigorous operational logic:
\begin{enumerate}
    \item \textbf{Expert Role and Context ($\mathcal{S}_{role}$):} Establishes the VLM's identity and enforces a 4-step reasoning chain: goal understanding, decomposition, primitive selection, and logical sequencing.
    \item \textbf{Action Primitive Library ($\mathcal{S}_{actions}$):} Provides the semantic mapping for set $\mathcal{A}$. It defines what an action does and \textit{how} it must be used safely. Actions are divided into general and specific measure conversion ones.
    \item \textbf{Composite Primitive Library ($\mathcal{S}_{composites}$):} Provides the semantic mapping for the execution logic in set $\mathcal{C}$.
    \item \textbf{Task Reasoning Rules ($\mathcal{S}_{reasoning\_rules}$, $\mathcal{S}_{BT\_rules}$):} Encodes high-level heuristics, such as restoring the environment to its original state and explaining unmeasurable derived parameters.
    \item \textbf{World Grounding and Metadata ($\mathcal{S}_{metadata}$):} Incorporates environmental metadata (e.g., object poses) and instructs the VLM to reconcile textual requests with visual evidence $\mathcal{I}$ to prevent spatial hallucinations.
    \item \textbf{Output Formalism ($\mathcal{S}_{output}$):} Defines the recursive list structure required for the \texttt{py\_trees} parser and provides few-shot examples of valid BT syntax.
\end{enumerate}

\subsubsection{\textbf{Variable Inputs (The Context)}}
The user provides only the variable context: the simulation description $\mathcal{D}$, the specific intent $\mathcal{R}$, and the visual state $\mathcal{I}$. By grounding the fixed grammar of $\mathcal{S}$ into the dynamic context of $(\mathcal{D}, \mathcal{R}, \mathcal{I})$, the framework achieves zero-shot task adaptation without requiring prompt engineering expertise from the end-user.

\begin{table}[]
\centering
\footnotesize
\caption{Composite Set $\mathcal{C}$ (Execution Logic)}
\label{tab:composites}
\begin{tabular}{@{}lll@{}}
\toprule
\textbf{Composite} & \textbf{Symbol} & \textbf{Execution Logic} \\ \midrule
\makecell{Sequence (with memory)} & $\{-\}$ & Executes children until one fails \\
Parallel & $/\_/$ & Executes all children concurrently \\ \bottomrule
\end{tabular}
\end{table}

\section{Experimental Setup}\label{sec: setup}
We validated the proposed Real2Sim framework through real-robot experiments involving contact-rich interactions and autonomous simulation construction. The system integrates the VLM-driven BT framework with a physics-based estimation layer and a MuJoCo simulation environment.

\subsection{Hardware, Software, and Asynchronous Execution}
Experiments utilize a 7-DoF Franka Emika Panda, controlled via ROS Noetic with a standard Cartesian impedance controller~\cite{Khatib} for compliant, task-agnostic probing, and an Intel RealSense D435i camera. The VLM reasoning engine utilizes either GPT-4o or Gemini 1.5 Pro (evaluated in Sec.~\ref{sec:evaluation}) to generate plans, which are subsequently parsed by the \texttt{py\_trees} library.

A critical operational advantage of using Behavior Trees over sequential scripts is their native capacity to manage asynchronous multi-modal data streams running at disparate frequencies (e.g., a 1 kHz robot control loop versus a 30 Hz camera feed). By evaluating conditions at a fixed tick-rate, the BT queries the latest cached sensor states without blocking the underlying real-time robot controllers. Crucially, sensor feedback is encapsulated directly within the atomic action primitives themselves. For example, a motion primitive like \texttt{MovePose} continuously evaluates the real-time Cartesian pose error against the cached sensor data, returning a \texttt{RUNNING} status to the tree until the target configuration is achieved. Furthermore, dedicated sensor data logging is strictly context-aware: the system only records and processes forces when the BT ticks a designated sensing leaf node (e.g., \texttt{MeasureForces}), ensuring computational efficiency and minimizing noise.

\subsection{Physical Parameter Estimation}

The targeted parameters $\Phi$ are estimated from the external wrench $\boldsymbol{w}_{\text{ext}}$, recovered from measured joint torques $\boldsymbol{\tau}$ via the transposed Jacobian pseudoinverse: $\boldsymbol{w}_{\text{ext}} = (\mathbf{J}(\boldsymbol{q})^\top)^{+} (\boldsymbol{\tau} - \boldsymbol{\tau}_{\text{gravity}} - \boldsymbol{\tau}_{\text{friction}})$. 

Mass $m$ is derived from the vertical force component $F_z$ during quasi-static lifting. For geometry acquisition, the \texttt{MoveUntilContact} primitive monitors the control loop until a torque threshold $\tau_c$ is breached, capturing the surface height $h$ from the end-effector's kinematics. To estimate friction coefficients ($\mu_s, \mu_d$), the BT parallelizes a planar pushing primitive with continuous force logging, computing $\mu = F_\parallel / F_z$ and distinguishing static from dynamic phases based on the motion state. Following BT execution, these derived values are autonomously injected into the MuJoCo \texttt{.xml} assets to complete the physics-aware digital twin $\mathcal{M}$.

\subsection{Simulation Environment Construction}
Following the successful execution of the BT, the estimated parameters are injected into the user's simulation template. The framework updates the MuJoCo \texttt{.xml} assets with the measured values. For geometric representation $\mathcal{G}$, the system utilizes simplified primitive shapes or convex hulls. These base shapes are scaled according to the VLM's initial visual bounding-box estimation and are subsequently verified by the robot's physical contact points, yielding a physics-aware digital twin ready for downstream algorithmic testing.

\section{Ablation Study and Quantitative Evaluation}\label{sec:evaluation}
To rigorously assess the framework's performance beyond qualitative case studies, we conducted a systematic evaluation focusing on structural reliability, generation repeatability, and the impact of specific prompt components. This quantitative analysis isolates the contribution of our engineered system prompt ($\mathcal{S}$) from the latent zero-shot capabilities of the foundational models.

\subsection{Ablation Study on Prompt Architecture}\label{subsec: ablation}
We evaluated the necessity of each input modality and system prompt module by selectively removing components from the $(\mathcal{S}, \mathcal{R}/\mathcal{D}, \mathcal{I})$ tuple. Each configuration was tested over 10 independent generation cycles using Gemini 1.5 Pro~\cite{team2023gemini}. 

\begin{table}[]
\centering
\footnotesize
\renewcommand{\arraystretch}{1.2}
\caption{Ablation Study Results (Success over 10 independent trials)}
\label{tab:ablation}
\begin{tabular}{@{}lcc@{}}
\toprule
\textbf{Configuration} & \textbf{Syntax Validity} & \textbf{Task Success}\\ \midrule
\textbf{Full Framework} & \textbf{10/10} & \textbf{10/10} \\ \midrule
No Visual Input ($\mathcal{I}$) & 10/10 & 5/10 \\ 
No Text Intent ($\mathcal{D}$ or $\mathcal{R}$) & 10/10 & -/10 \\ \midrule
No Role Prompt ($\mathcal{S}_{role}$) & 10/10 & 10/10 \\ 
No Action Lib ($\mathcal{S}_{actions}$) & 10/10 & 0/10 \\ 
No Composite Lib ($\mathcal{S}_{composites}$) & 10/10 & 8/10 \\ 
\makecell[l]{No Reasoning Rules\\($\mathcal{S}_{reasoning\_rules}$, $\mathcal{S}_{BT\_rules}$)} & 10/10 & 3/10 \\ 
No Output Few-Shot ($\mathcal{S}_{output}$) & 0/10 & 0/10 \\ \midrule
\textbf{No System Prompt $\mathcal{S}$ (Baseline)} & 0/10 & 0/10 \\ \bottomrule
\end{tabular}
\end{table}

We measure two distinct metrics:
\begin{enumerate}
    \item \textbf{Syntax Validity:} The number of trials (out of 10) where the VLM generated a perfectly formatted, parseable structure that compiled into a valid \texttt{py\_trees} object without manual correction.
    \item \textbf{Task Success:} The number of trials where the generated BT successfully orchestrated the necessary sequence to acquire the exact missing parameter set $\Phi$. Success is achieved if and only if the executed BT triggers the correct sensing primitives on the correct target objects without triggering safety aborts or redundant measurements. This is evaluated purely on the resulting execution trace, independent of human grading.
\end{enumerate}

As shown in Tab.~\ref{tab:ablation}, removing modules like $\mathcal{S}_{actions}$ or $\mathcal{S}_{output}$ causes catastrophic drops in Syntax Validity, proving that the structural rigor of the framework is strictly driven by the prompt architecture rather than the VLM's inherent robotics knowledge. Similarly, omitting visual context ($\mathcal{I}$) severely degrades Task Success due to spatial hallucinations when spatial reasoning is needed.

\subsection{Generation Repeatability and Model Agnosticism}
A primary concern with generative planning is stochasticity and outcome uncertainty. To verify that our Semantic Task Decomposition relies on the mathematical structure of the prompt rather than overfitting to a specific VLM, we tested generation repeatability across two state-of-the-art models: GPT-4o~\cite{openai2024gpt4technicalreport} and Gemini 1.5 Pro. 

Each model was tasked with generating the Real2Sim BT for the identical, feasible, target object across $N=20$ trials. As demonstrated in Tab.~\ref{tab:repeatability}, both models maintain high structural integrity. This indicates that the strict formalism of $\mathcal{S}$ effectively constrains the LLM's output manifold, acting as a stabilizing filter against typical autoregressive hallucinations.

\begin{table}[h]
\centering
\footnotesize
\renewcommand{\arraystretch}{1.2}
\caption{Model Agnosticism and Repeatability ($N=20$ trials)}
\label{tab:repeatability}
\begin{tabular}{@{}lcc@{}}
\toprule
\textbf{Foundational Model} & \textbf{Syntax Validity} & \textbf{Task Success}\\ \midrule
GPT-4 & 20/20 & 20/20 \\ 
Gemini 1.5 Pro & 20/20 & 20/20 \\ \bottomrule
\end{tabular}
\end{table}

\section{Real-World Experimental Results and Discussion}\label{sec:experiments}
We evaluate the proposed framework through a series of real-robot experiments designed to assess its ability to translate high-level intent into physically grounded interaction policies.

The experiments are conducted on a torque-controlled Franka Emika Panda robot. Throughout all scenarios, the action set $\mathcal{A}$, the composite set $\mathcal{C}$, and the system prompt $\mathcal{S}$ remain constant, demonstrating the zero-shot generalization. 

\subsection{Task-Specific Parameter Estimation}
The first set of experiments validates the basic Real2Sim pipeline, where the VLM must identify missing physical properties from a user request $\mathcal{R}$ or an incomplete simulation description $\mathcal{D}$.

\subsubsection{\textbf{Estimation of Mass and Geometry}} 
In this scenario, the user provides a natural language request $\mathcal{R}$: \textit{"We want to build a simulation, but we don't know the mass of the bottle and the height of the table."} As shown in Fig.~\ref{fig:BT} (left), the VLM autonomously sequences surface probing and bottle weighing. Recognizing the quasi-static context, it actively prunes irrelevant conversion actions and discards physically irrelevant conversion actions from the vocabulary (e.g., \texttt{Force2Velocity}, \texttt{Force2Momentum}, \texttt{Pose2Plan}), strictly selecting \texttt{Pose2Height} and \texttt{Force2Mass} to process the raw sensor buffers.
\begin{figure}[]
    \centering
    
    \begin{tikzpicture}[placeholder/.style={draw=gray!30, fill=white, rounded corners=1pt, font=\fontsize{6}{7}\selectfont, align=center},
    arrow/.style={-{Stealth[scale=0.8]}, thick, draw=gray!60},
    % Group Styles
    group_bg/.style={draw=gray!15, fill=#1, rounded corners=3pt, inner sep=3pt}]

    \node[placeholder, right=0.05cm of bt, xshift=0.1cm, 
      font=\fontsize{5}{6}\selectfont\ttfamily, 
      inner sep=2pt, align=left, draw=unipdred!40] (thumb_bt) {
    \textbf{Real2Sim Task Plan:}\\
    \btS{\{-\}} Sequence\\
    \hspace{10pt}\btA{$\rightarrow$} MoveJoints(home\_config)\\
    \hspace{10pt}\btA{$\rightarrow$} OpenGripper()\\
    \hspace{10pt}\btA{$\rightarrow$} MovePose(table\_approach+z)\\
    \hspace{10pt}\btS{\{-\}} Sequence\\
    \hspace{15pt}\btA{$\rightarrow$} MoveUntilContact(-z)\\
    \hspace{15pt}\btA{$\rightarrow$} \textbf{MeasureGripperPose()}\\
    \hspace{15pt}\btA{$\rightarrow$} \textbf{Pose2Height(gripper\_pose)}\\
    \hspace{10pt}\btA{$\rightarrow$} MovePose(table\_approach+z)\\
    \hspace{10pt}\btA{$\rightarrow$} MovePose(bottle\_pose+z)\\
    \hspace{10pt}\btA{$\rightarrow$} MovePose(bottle\_pose)\\
    \hspace{10pt}\btA{$\rightarrow$} CloseGripper()\\
    \hspace{10pt}\btA{$\rightarrow$} MovePose(bottle\_pose+z)\\
    \hspace{10pt}\btS{\{-\}} Sequence\\
    \hspace{15pt}\btA{$\rightarrow$} \textbf{MeasureForces()}\\
    \hspace{15pt}\btA{$\rightarrow$} \textbf{Force2Mass(gripper\_forces)}\\
    \hspace{10pt}\btA{$\rightarrow$} MovePose(bottle\_pose)\\
    \hspace{10pt}\btA{$\rightarrow$} OpenGripper()\\
    \hspace{10pt}\btA{$\rightarrow$} MovePose(bottle\_pose+z)\\
    \hspace{10pt}\btA{$\rightarrow$} MoveJoints(home\_config)
};
\node[placeholder, right=0.05cm of thumb_bt, xshift=0.1cm, 
      font=\fontsize{5}{6}\selectfont\ttfamily, 
      inner sep=2pt, align=left, draw=unipdred!40] (thumb_bt1) {
    \textbf{Real2Sim Task Plan:}\\
    \btS{\{-\}} Sequence\\
    \hspace{10pt}\btA{$\rightarrow$} MoveJoints(home\_config)\\
    \hspace{10pt}\btA{$\rightarrow$} OpenGripper()\\
    \hspace{10pt}\btA{$\rightarrow$} MovePose(green\_bottle\_pose+z\_off)\\
    \hspace{10pt}\btA{$\rightarrow$} MovePose(green\_bottle\_pose)\\
    \hspace{10pt}\btA{$\rightarrow$} CloseGripper()\\
    \hspace{10pt}\btA{$\rightarrow$} MovePose(green\_bottle\_pose+z\_off)\\
    \hspace{10pt}\btS{\{-\}} Sequence\\
    \hspace{15pt}\btA{$\rightarrow$} \textbf{MeasureForces()}\\
    \hspace{15pt}\btA{$\rightarrow$} \textbf{Force2Mass(gripper\_forces)}\\
    \hspace{10pt}\btA{$\rightarrow$} MovePose(green\_bottle\_pose)\\
    \hspace{10pt}\btA{$\rightarrow$} OpenGripper()\\
    \hspace{10pt}\btA{$\rightarrow$} MovePose(bottle\_pose+z\_off)\\
    \hspace{10pt}\btA{$\rightarrow$} MoveJoints(home\_config)
    
};
        
    \end{tikzpicture}
    \caption{Example of BT generated for the estimation of table height and mass of the blue bottle (left). The VLM uses a sequence node to ensure table height is acquired before manipulation. Example of BT generated for the estimation of the mass of the green bottle (right).}
    \label{fig:BT}
\end{figure}

\begin{figure}[h]
    \centering
    \includegraphics[width=0.24\linewidth]{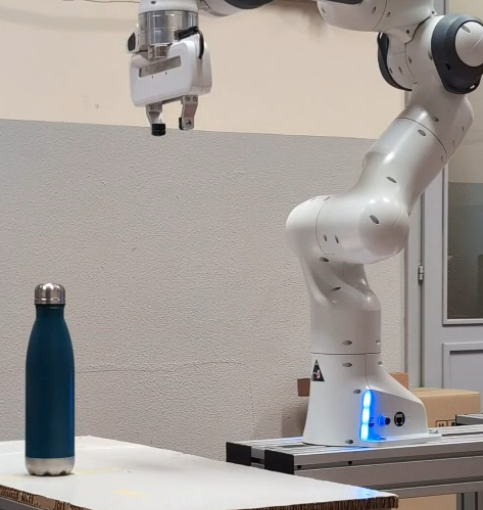}
    \includegraphics[width=0.24\linewidth]{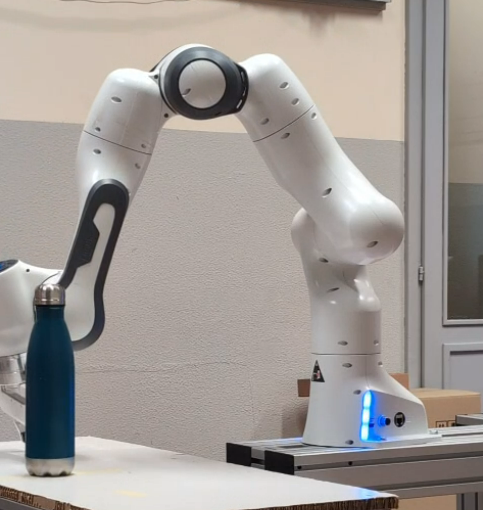}
    \includegraphics[width=0.24\linewidth]{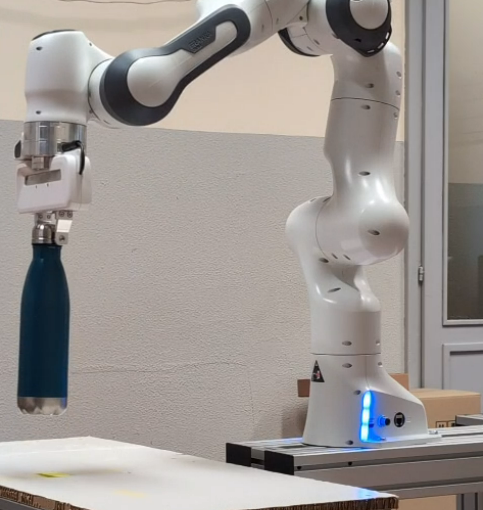}
    \includegraphics[width=0.24\linewidth]{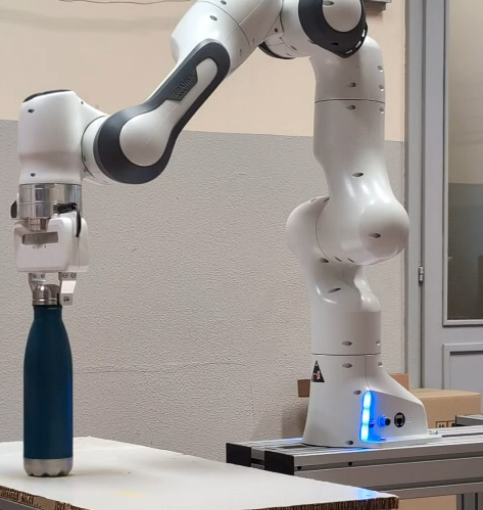}
    \caption{Execution sequence of table probing followed by bottle weighing.}
    \label{fig: scenario 1 sequence}
\end{figure}

The execution sequence is illustrated in Fig.~\ref{fig: scenario 1 sequence}. Quantitative results, summarized in Tab.~\ref{tab:combined_results}, show that the estimated height and mass are highly consistent with manual ground-truth measurements, with the robot achieving sub-millimeter precision in height detection through compliant contact.

\subsubsection{\textbf{Intent-Driven Selective Acquisition}}\label{Sec: intent driven}
To test the grounding of the simulation description $\mathcal{D}$, we provided an incomplete MuJoCo \texttt{.xml} file. When $\mathcal{D}$ lacked only the mass of a specific object, the VLM generated a pruned BT (Fig.~\ref{fig:BT} right) that skipped the table-probing phase and other objects testing. 

Furthermore, in scenes with multiple objects, the model correctly identified and interacted only with the objects missing in $\mathcal{D}$ (or potentially referenced in $\mathcal{R}$), avoiding redundant interactions common in task-agnostic pipelines.

\subsubsection{\textbf{Cross-Domain Generalization (Synthetic to Real)}}
We further evaluated the framework using synthetic images generated from MuJoCo. The VLM produced identical interaction strategies for both synthetic and real inputs, suggesting that the reasoning logic is robust to the domain gap between simulation and reality, provided the semantic content is preserved.

\subsubsection{\textbf{Object Occlusion and Scene Clearing}}
\begin{figure}[]
    \centering
    \includegraphics[width=0.24\linewidth]{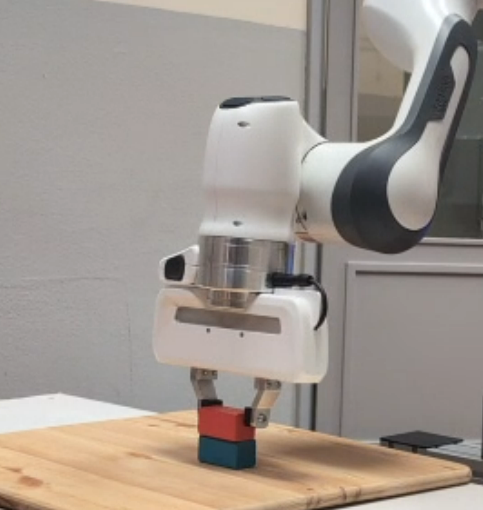}
    \includegraphics[width=0.24\linewidth]{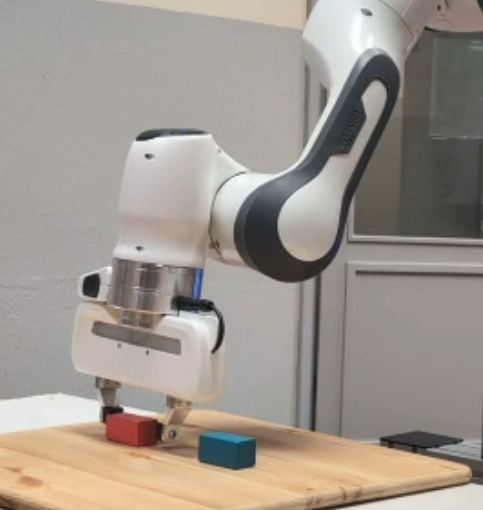}
    \includegraphics[width=0.24\linewidth]{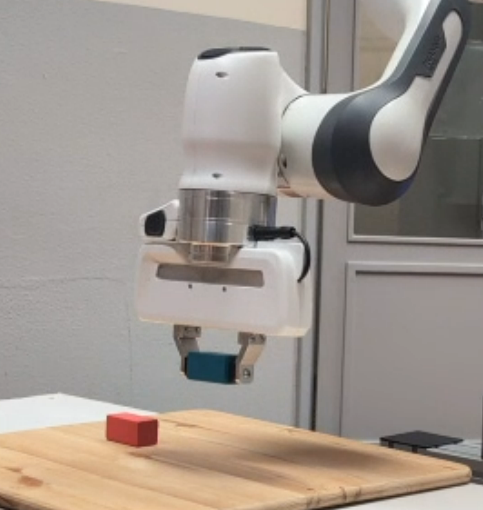}
    \includegraphics[width=0.24\linewidth]{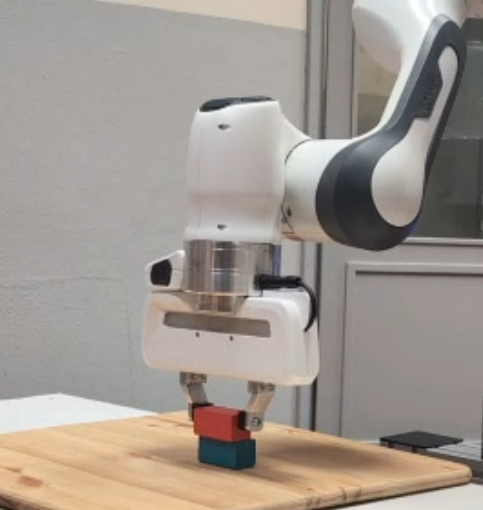}
    \caption{Clearing occlusions: The robot moves the red block to access the blue target.}
    \label{fig: boxes sequence}
\end{figure}

As shown in Fig.~\ref{fig: boxes sequence}, a target object $o_1$ (blue box) is occluded by $o_2$ (red box). Without explicit instruction to move $o_2$, the VLM inferred from the visual context $\mathcal{I}$ that $o_2$ must be displaced to access $o_1$. The generated BT successfully moved the occluding object to a temporary pose $x_1$, performed the measurement (Tab.~\ref{tab:combined_results}), and restored the original scene configuration.

\subsubsection{\textbf{Derived Parameter Estimation (Friction)}}
This scenario required the estimation of friction coefficients $\mu_s, \mu_d$. Because there is no explicit \texttt{Force2Friction} action in $\mathcal{A}$, the VLM was forced to reason about the underlying physics. Demonstrating its compositional ability, the model actively discarded the unrelated conversion nodes (such as \texttt{Force2Velocity} or \texttt{Force2Momentum}). Instead, it generated a complex interaction (Fig.~\ref{fig: BT3}) that parallelized weighing the object with a lateral sliding motion, utilizing raw data logging via \texttt{MeasureForces} to capture the tangential shear forces required for the user to derive the coefficient manually.
\begin{figure}[]
    \centering
    
    \begin{tikzpicture}[placeholder/.style={draw=gray!30, fill=white, rounded corners=1pt, font=\fontsize{6}{7}\selectfont, align=center},
    arrow/.style={-{Stealth[scale=0.8]}, thick, draw=gray!60},
    % Group Styles
    group_bg/.style={draw=gray!15, fill=#1, rounded corners=3pt, inner sep=3pt}]

    \node[placeholder, right=0.05cm of bt, xshift=0.1cm, 
      font=\fontsize{5}{6}\selectfont\ttfamily, 
      inner sep=2pt, align=left, draw=unipdred!40] (thumb_bt) {
    \textbf{Friction \& Mass Strategy:}\\
    \btS{\{-\}} Sequence\\
    \hspace{10pt}\btA{$\rightarrow$} MoveJoints(home\_config)\\
    \hspace{10pt}\btA{$\rightarrow$} OpenGripper()\\
    \hspace{10pt}\btA{$\rightarrow$} MovePose(bottle\_pose+z\_off)\\
    \hspace{10pt}\btA{$\rightarrow$} MovePose(bottle\_pose)\\
    \hspace{10pt}\btA{$\rightarrow$} CloseGripper()\\
    \hspace{10pt}\btA{$\rightarrow$} MovePose(bottle\_pose+z\_off)\\
    \hspace{10pt}\btS{\{-\}} Sequence\\
    \hspace{15pt}\btA{$\rightarrow$} \textbf{MeasureForces()}\\
    \hspace{15pt}\btA{$\rightarrow$} \textbf{Force2Mass(gripper\_forces)}\\
    \hspace{10pt}\btA{$\rightarrow$} MovePose(bottle\_pose)\\
    \hspace{10pt}\btS{/\_/} Parallel\\
    \hspace{20pt}\btA{$\rightarrow$} MovePose(table\_approach)\\
    \hspace{20pt}\btA{$\rightarrow$} \textbf{MeasureForces()}\\
    \hspace{10pt}\btA{$\rightarrow$} OpenGripper()\\
    \hspace{10pt}\btA{$\rightarrow$} MovePose(table\_approach+z\_off)\\
    \hspace{10pt}\btA{$\rightarrow$} MoveJoints(home\_config)\\

    \textbf{output\_text}="The mass of the blue bottle and the sliding force across the\\  table surface have been captured. To derive the coefficient of friction\\, divide the average horizontal force measured during the
    sliding MovePose\\ by the vertical force (gravity × mass) measured 
    during the lift."
};
        
    \end{tikzpicture}
    \caption{Behavior Tree generated for the simultaneous estimation of mass and friction coefficients. The VLM parallelizes a sliding Cartesian motion with real-time force logging to acquire the necessary data for friction estimation.}
    \label{fig: BT3}
\end{figure}
The explanation text $\mathcal{E}$ provided by the model confirmed its strategy to provide the user with sufficient raw data to compute the derived friction values (Tab.~\ref{tab:combined_results}).

\begin{table}[]
\centering
\footnotesize
\caption{Real-World Parameter Estimation Results across Scenarios}
\label{tab:combined_results}
\begin{tabular}{@{}llcc@{}}
\toprule
\textbf{Scenario} & \textbf{Parameter} & \makecell{\textbf{Estimated} \\(Mean $\pm$ SD)} & \textbf{Ground Truth} \\ \midrule
1. Basic Setup & Bottle Mass [kg] & $0.257 \pm 0.022$ & $0.254$ \\
1. Basic Setup & Table Height [m] & $0.7642 \pm 0.00004$ & $0.765$ \\
2. Friction & Coeff. ($\mu_s$ / $\mu_d$) & \makecell{$0.41 \pm 0.11$\\ $0.34 \pm 0.20$} & N/A \\
3. Occlusion & Blue Mass [kg] & $0.022 \pm 0.015$ & $0.016$ \\ \bottomrule
\end{tabular}
\end{table}
\subsection{Baseline Comparison: Intent-Driven vs. Exhaustive BT}
To quantitatively demonstrate the operational efficiency of Semantic Task Decomposition, we compared the VLM-generated Behavior Trees against a standard Exhaustive Parameter Identification baseline. This baseline represents a task-agnostic Real2Sim BT: it iterates through all detected objects $\mathcal{O}$ and executes the full suite of sensing primitives $\mathcal{A}_{\text{sense}}$ to populate a complete physical parameter matrix. We evaluated both systems on a tabletop scene containing three distinct objects (Sec.~\ref{Sec: intent driven}), with a user intent $\mathcal{R}$ requesting only the mass of a single specific object. 

\begin{itemize}
    \item \textbf{Exhaustive Baseline:} Executed $N_{total} = 29$ atomic actions, requiring $\simeq 62$ seconds to probe heights, weigh, and test friction for all three objects.
    \item \textbf{Intent-Driven VLM:} Generated a pruned BT executing only $N_{vlm} = 12$ atomic actions, requiring $\simeq 33$ seconds. 
\end{itemize}

By actively pruning task-irrelevant interactions, the proposed framework achieved an operational efficiency gain of $47\%$, minimizing unnecessary physical wear on the robot and significantly accelerating the digital twin construction.

\subsection{Advanced Generalization Scenarios}

A common limitation of traditional active system identification is an over-reliance on rigid, predefined heuristics. To demonstrate that our framework acts as a generalized semantic optimizer rather than a pre-scripted sequence, we tested it on two advanced boundary scenarios.

\subsubsection{Hidden Geometry (Physical Overrides Visual)}
We introduced a scenario where the visual estimation of the table height $\mathcal{I}$ was intentionally corrupted by placing a thick, visually deceptive cloth over the workspace. Relying solely on vision models would yield an incorrect collision geometry for the MuJoCo simulation. Guided by the system rules, the VLM generated a BT utilizing the \texttt{MoveUntilContact} primitive. The robot compliant impedance controller safely compressed the cloth, registering the $\tau_c$ threshold only when hitting the rigid table beneath. This successfully decoupled the true physical geometry from the visual artifact, proving the necessity of contact-rich active perception in Real2Sim pipelines.

\subsubsection{Active Heuristic Optimization (Irregular Objects)}
To verify that the VLM can construct novel exploratory behaviors, the system was tasked with estimating the mass of an irregular, highly asymmetric object.
\begin{figure}[]
    \centering
    
    \begin{tikzpicture}[placeholder/.style={draw=gray!30, fill=white, rounded corners=1pt, font=\fontsize{6}{7}\selectfont, align=center},
    arrow/.style={-{Stealth[scale=0.8]}, thick, draw=gray!60},
    % Group Styles
    group_bg/.style={draw=gray!15, fill=#1, rounded corners=3pt, inner sep=3pt}]

   \node[placeholder, right=0.05cm of bt, xshift=0.1cm, 
      font=\fontsize{5}{6}\selectfont\ttfamily, 
      inner sep=2pt, align=left, draw=unipdred!40] (thumb_bt) {
    \textbf{Heuristic CoG Discovery:}\\
    \btS{\{-\}} Sequence\\
    \hspace{10pt}\btA{$\rightarrow$} MoveJoints(home\_config)\\
    \hspace{10pt}\btA{$\rightarrow$} OpenGripper()\\
    \hspace{10pt}\btS{\{-\}} Sequence\\
    \hspace{20pt}\btA{$\rightarrow$} MovePose(hammer+z)\\
    \hspace{20pt}\btA{$\rightarrow$} MovePose(hammer)\\
    \hspace{20pt}\btA{$\rightarrow$} CloseGripper()\\
    \hspace{20pt}\btA{$\rightarrow$} MovePose(hammer+z)\\
    \hspace{20pt}\btA{$\rightarrow$} \textbf{MeasureForces()}\\
    \hspace{20pt}\btA{$\rightarrow$} MovePose(hammer)\\
    \hspace{20pt}\btA{$\rightarrow$} OpenGripper()\\
    \hspace{10pt}\btS{\{-\}} Sequence\\
    \hspace{20pt}Measurement sequence for hammer+ 1*x\_offset...\\
    \hspace{10pt}\btS{\{-\}} Sequence\\
    \hspace{20pt}Measurement sequence for hammer- 1*x\_offset...\\
    \hspace{10pt}\btA{$\rightarrow$} MoveJoints(home\_config)

\\\textbf{output\_text}="The exploration sequence captured the contact height of the hammer\\ and recorded the mass and torque at three distinct grasping points along the handle\\ (x\_offset * 1 and x\_offset * -1). To derive the Center of Mass (CoM), the\\ user must analyze the distribution of the measured forces relative to the grasp\\ positions; the CoM lies at the point where the static torque about the\\ gripper's Y-axis would be zero."
};
        
    \end{tikzpicture}
    \caption{Example of BT generated for the estimation of the Center of mass of a hammer. The VLM uses x offsets to pick it up from different positions.}
    \label{fig: BT4}
\end{figure}

Without task-specific reprogramming, the VLM iteratively adapted its strategy. It generated a BT loop that grasped the object at 3 different candidate poses, using the \texttt{MeasureForces} primitive to evaluate the $\tau_x, \tau_y$ torque vectors. The VLM decided to test 3 new poses, without specific instructions. In the prompt, the model was only instructed to symbolically scale the pose of an object with offsets with scaling factors (e.g. \texttt{$\alpha$ * x\_offset}).
Notably, the VLM correctly discarded the \texttt{Force2Mass} conversion action during the offset grasps, recognizing that a mass calculation is physically invalid while the object is undergoing severe torsional imbalance. The BT successfully identified the center grasp pose that minimized torsional imbalance, allowing for an accurate quasi-static mass estimation using \texttt{Force2Mass}. This behavior proves the VLM functions as an active heuristic optimizer, intelligently pruning and sequencing atomic primitives to solve complex physical constraints.

\section{Boundaries of Autonomy and Failure Analysis}\label{sec:limits}
While the framework successfully automates parameter acquisition, extended testing revealed distinct boundary conditions related to hardware constraints and VLM generation.

\subsection{Semantic Hallucination and Tool Use}
Unconstrained LLMs are notoriously prone to vocabulary hallucination for out-of-bounds requests. Queried with an unmeasurable property (\textit{"measure water temperature"}), the VLM correctly evaluated the limits of $\mathcal{A}$. It generated a safe, idle Behavior Tree alongside an explicit semantic rejection in $\mathcal{E}$: \textit{"The requested parameter cannot be measured..."} This proves the system enforces strict semantic boundaries; extending the framework to novel parameters fundamentally requires expanding the sensing nodes in $\mathcal{A}$.

To test external tool adaptability, we provided the pose of a thermometer. The VLM successfully synthesized a kinematic tool-use sequence: navigating, grasping the tool, probing the bottles, and returning it. However, it intelligently decoupled kinematics from data ingestion, noting: \textit{"Temperature values must be read manually... the Action List does not include direct data ingestion"}. Thus, while movement compositionality seamlessly extends to external tools, fully autonomous perception remains bounded by integrated digital sensing primitives. 
Crucially, when the framework encounters an unsolvable task, it does not resort to unpredictable behavior, and the system mitigates outcome uncertainty. If the VLM determines that the required parameter $\Phi$ cannot be isolated using the available primitives in $\mathcal{A}$, it halts BT generation and utilizes the explanation variable $\mathcal{E}$ to inform the user of the physical impossibility (Fig.~\ref{fig: BT4}). If the VLM hallucinates a geometrically unreachable pose or if a grasp slips during execution, the Behavior Tree acts as a deterministic safety filter (Sec.~\ref{Sec: scene undertsanding}).

While the proposed framework successfully automates parameter acquisition for tabletop scenes, extended testing revealed distinct boundary conditions related to hardware constraints and algorithmic limitations. The system is fundamentally bounded by the robot's physical capabilities and the vocabulary of $\mathcal{A}$ for direct parameter acquisition (transformation actions in Tab.~\ref{tab:actions}), while also being able to tell the user which measurements to read to retrieve them.
Therefore, while LLM uncertainty is unavoidable, the reactive hierarchy of the Behavior Tree successfully transforms potential generative failures into safe, controlled hardware aborts.

\subsection{Syntactic Breakdown and Safety Filtering}
To test combinatorial scaling limits, the VLM was instructed to parallelize optimization in a 15-object scene. Under the baseline prompt, while the model surprisingly maintained complex syntactic parsing, it initially failed in physical resource reasoning by wrapping 15 distinct spatial manipulation sequences inside \texttt{Parallel} nodes. However, explicitly updating the system prompt to define the single-arm hardware constraint immediately resolved this issue. This demonstrates that while textual models may default to unconstrained parallelization, they successfully map algorithmic logic to physical embodiment when provided proper grounding. Furthermore, even if such resource-blind hallucinations occur, the Behavior Tree architecture inherently mitigates these uncertainties through reactive hardware filtering. The first executing motion primitive claims the controller interface, safely blocking physically impossible parallel branches from executing.

\subsection{Hardware and Metrological Limits}
The system's estimation accuracy is fundamentally bounded by the robot's physical payload and proprioceptive noise, restricting dynamic two-handed manipulation. Furthermore, friction estimation ($\mu_s, \mu_d$) exhibits high variance due to resolving delicate planar contact forces through the geometric Jacobian, which is subject to internal joint friction. However, because our core contribution is the \textit{methodology of intent-driven semantic sequencing}, future implementations can seamlessly achieve tighter confidence intervals by simply integrating a dedicated wrist force/torque sensor without requiring any alterations to the underlying VLM reasoning architecture.

\section{Conclusions and Future Work}\label{sec:conclusion}
We presented an autonomous, intent-driven Real2Sim framework that leverages Semantic Task Decomposition to translate high-level user requests into executable Behavior Trees (BTs). Unlike exhaustive, task-agnostic pipelines, our VLM-powered approach selectively estimates missing physical parameters using a vocabulary of atomic robotic primitives. Real-world experiments on a torque-controlled manipulator demonstrated estimation accuracy and operational efficiency gains over baseline methods. Crucially, the BT's reactive hierarchy acts as a deterministic safety filter, successfully mitigating generative VLM hallucinations, such as semantic overreach and physical resource blindness, during interactions. 

While highly effective for rigid-body tabletop scenes, the framework's autonomy remains bounded by the rigidity of its sensing vocabulary and the inherent limitations of foundational models in mapping algorithmic parallelization to physical embodiment constraints. Future work will relax pose metadata assumptions via vision-based 6D pose estimation~\cite{wen2024foundationposeunified6dpose}, integrate visuo-tactile sensing for deformable objects, and explore multi-agent LLM verification to resolve combinatorial scaling limits in highly parallelized tasks.

\bibliographystyle{IEEEtran}

\bibliography{references}

@inproceedings{estobjinertia,
author = {Mavrakis, Nikos and Ghalamzan, Amir and Stolkin, Rustam},
booktitle={2020 IEEE/RSJ International Conference on Intelligent Robots and Systems},
year = {2020},
month = {07},
pages = {},
title = {Estimating An Object's Inertial Parameters By Robotic Pushing: A Data-Driven Approach},
doi = {10.1109/IROS45743.2020.9341112}
}

@inproceedings{todorov2012mujoco,
  title={MuJoCo: A physics engine for model-based control},
  author={Todorov, Emanuel and Erez, Tom and Tassa, Yuval},
  booktitle={2012 IEEE/RSJ International Conference on Intelligent Robots and Systems},
  pages={5026--5033},
  year={2012},
  organization={IEEE},
  doi={10.1109/IROS.2012.6386109}
}

@article{ROSEN2015567,
title = {About The Importance of Autonomy and Digital Twins for the Future of Manufacturing},
journal = {IFAC-PapersOnLine},
volume = {48},
number = {3},
pages = {567-572},
year = {2015},
note = {15th IFAC Symposium onInformation Control Problems inManufacturing},
issn = {2405-8963},
doi = {https://doi.org/10.1016/j.ifacol.2015.06.141},
url = {},
author = {Roland Rosen and Georg {von Wichert} and George Lo and Kurt D. Bettenhausen}
}

@article{pfaff2025scalablereal2simphysicsawareasset,
      title={Scalable Real2Sim: Physics-Aware Asset Generation Via Robotic Pick-and-Place Setups}, 
      author={Nicholas Pfaff and Evelyn Fu and Jeremy Binagia and Phillip Isola and Russ Tedrake},
      year={2025},
      eprint={2503.00370},
      archivePrefix={arXiv},
      primaryClass={cs.RO},
      url={https://arxiv.org/abs/2503.00370}, 
}

@article{heiden2019real2sim,
  author  = {Eric Heiden and David Millard and Gaurav S. Sukhatme},
  title   = {Real2Sim Transfer using Differentiable Physics},
  journal = {R:SS Workshop on Closing the Reality Gap in Sim2real Transfer for Robotic Manipulation},
  year    = {2019}
}

@article{tao2019,
author = {Tao, F. and et al},
year = {2019},
month = {01},
pages = {1-18},
title = {Five-dimension digital twin model and its ten applications},
volume = {25},
journal = {Jisuanji Jicheng Zhizao Xitong/Computer Integrated Manufacturing Systems, CIMS},
doi = {10.13196/j.cims.2019.01.001}
}

@misc{Colledanchise_2018,
   title={Behavior Trees in Robotics and AI},
   ISBN={9780429950902},
   url={http://dx.doi.org/10.1201/9780429489105},
   DOI={10.1201/9780429489105},
   publisher={CRC Press},
   author={Colledanchise, Michele and Ögren, Petter},
   year={2018},
   month=jul }

@article{kim24openvla,
    title={OpenVLA: An Open-Source Vision-Language-Action Model},
    author={{Moo Jin} Kim and et al},
    journal = {arXiv preprint arXiv:2406.09246},
    year={2024}
}

@inproceedings{driess2023palme,
    title={PaLM-E: An Embodied Multimodal Language Model},
    author={Driess, Driess and et al},
    booktitle={arXiv preprint arXiv:2303.03378},
    year={2023}
}

@misc{brohan2023rt2visionlanguageactionmodelstransfer,
      title={RT-2: Vision-Language-Action Models Transfer Web Knowledge to Robotic Control}, 
      author={Anthony, Brohan and et al},
      year={2023},
      eprint={2307.15818},
      archivePrefix={arXiv},
      primaryClass={cs.RO},
      url={https://arxiv.org/abs/2307.15818}, 
}

@article{IOVINO2022104096,
title = {A survey of Behavior Trees in robotics and AI},
journal = {Robotics and Autonomous Systems},
volume = {154},
pages = {104096},
year = {2022},
issn = {0921-8890},
doi = {https://doi.org/10.1016/j.robot.2022.104096},
author = {Matteo Iovino and Edvards Scukins and Jonathan Styrud and Petter Ögren and Christian Smith}
}

@misc{wake2025vlm-driven,
author = {Wake, Naoki and et al},
title = {VLM-driven Behavior Tree for Context-aware Task Planning},
howpublished = {arXiv},
year = {2025},
month = {January},
url = {https://www.microsoft.com/en-us/research/publication/vlm-driven-behavior-tree-for-context-aware-task-planning/},
}

@misc{lykov2023llmbrainaidrivenfastgeneration,
      title={LLM-BRAIn: AI-driven Fast Generation of Robot Behaviour Tree based on Large Language Model}, 
      author={Artem Lykov and Dzmitry Tsetserukou},
      year={2023},
      eprint={2305.19352},
      archivePrefix={arXiv},
      primaryClass={cs.RO},
      url={https://arxiv.org/abs/2305.19352}, 
}

@misc{wen2024foundationposeunified6dpose,
      title={FoundationPose: Unified 6D Pose Estimation and Tracking of Novel Objects}, 
      author={Bowen Wen and Wei Yang and Jan Kautz and Stan Birchfield},
      year={2024},
      eprint={2312.08344},
      archivePrefix={arXiv},
      primaryClass={cs.CV},
      url={https://arxiv.org/abs/2312.08344}, 
}

@misc{ahmad2025unifiedframeworkrealtimefailure,
      title={A Unified Framework for Real-Time Failure Handling in Robotics Using Vision-Language Models, Reactive Planner and Behavior Trees}, 
      author={Faseeh Ahmad and et al},
      year={2025},
      eprint={2503.15202},
      archivePrefix={arXiv},
      primaryClass={cs.RO},
      url={https://arxiv.org/abs/2503.15202}, 
}

@ARTICLE{Khatib,
  author={Khatib, O.},
  journal={IEEE Journal on Robotics and Automation}, 
  title={A unified approach for motion and force control of robot manipulators: The operational space formulation}, 
  year={1987},
  volume={3},
  number={1},
  pages={43-53},
  doi={10.1109/JRA.1987.1087068}}

@misc{ling2025scenethesislanguagevisionagentic,
      title={Scenethesis: A Language and Vision Agentic Framework for 3D Scene Generation}, 
      author={Lu Ling and et al},
      year={2025},
      eprint={2505.02836},
      archivePrefix={arXiv},
      primaryClass={cs.CV},
      url={https://arxiv.org/abs/2505.02836}, 
}

@misc{liu2025agentic3dscenegeneration,
      title={Agentic 3D Scene Generation with Spatially Contextualized VLMs}, 
      author={Xinhang Liu and Yu-Wing Tai and Chi-Keung Tang},
      year={2025},
      eprint={2505.20129},
      archivePrefix={arXiv},
      primaryClass={cs.CV},
      url={https://arxiv.org/abs/2505.20129}, 
}

@misc{wang2025embodiedgengenerative3dworld,
      title={EmbodiedGen: Towards a Generative 3D World Engine for Embodied Intelligence}, 
      author={Xinjie Wang and et al},
      year={2025},
      eprint={2506.10600},
      archivePrefix={arXiv},
      primaryClass={cs.RO},
      url={https://arxiv.org/abs/2506.10600}, 
}

@article{Kim_2025,
   title={ASBI: Leveraging informative real-world data for active black-box simulator tuning},
   volume={55},
   ISSN={1573-7497},
   url={http://dx.doi.org/10.1007/s10489-025-06934-z},
   DOI={10.1007/s10489-025-06934-z},
   number={16},
   journal={Applied Intelligence},
   publisher={Springer Science and Business Media LLC},
   author={Kim, Gahee and Matsubara, Takamitsu},
   year={2025},
   month=oct }

@misc{memmel2024asidactiveexplorationidentification,
      title={ASID: Active Exploration for System Identification in Robotic Manipulation}, 
      author={Marius Memmel and Andrew Wagenmaker and Chuning Zhu and Patrick Yin and Dieter Fox and Abhishek Gupta},
      year={2024},
      eprint={2404.12308},
      archivePrefix={arXiv},
      primaryClass={cs.RO},
      url={https://arxiv.org/abs/2404.12308}, 
}

@INPROCEEDINGS{9341401,
  author={Possas, Rafael and Barcelos, Lucas and Oliveira, Rafael and Fox, Dieter and Ramos, Fabio},
  booktitle={2020 IEEE/RSJ International Conference on Intelligent Robots and Systems (IROS)}, 
  title={Online BayesSim for Combined Simulator Parameter Inference and Policy Improvement}, 
  year={2020},
  volume={},
  number={},
  pages={5445-5452},
  doi={10.1109/IROS45743.2020.9341401}}

@misc{margolis2023learningphysicalpropertiesactive,
      title={Learning to See Physical Properties with Active Sensing Motor Policies}, 
      author={Gabriel B. Margolis and Xiang Fu and Yandong Ji and Pulkit Agrawal},
      year={2023},
      eprint={2311.01405},
      archivePrefix={arXiv},
      primaryClass={cs.RO},
      url={https://arxiv.org/abs/2311.01405}, 
}

@article{team2023gemini,
  title={Gemini: a family of highly capable multimodal models},
  author={Team, Gemini and Anil, Rohan and Borgeaud, Sebastian and Alayrac, Jean-Baptiste and Yu, Jiahui and Soricut, Radu and Schalkwyk, Johan and Dai, Andrew M and Hauth, Anja and Millican, Katie and others},
  journal={arXiv preprint arXiv:2312.11805},
  year={2023}
}

@misc{openai2024gpt4technicalreport,
      title={GPT-4 Technical Report}, 
      author={OpenAI},
      year={2024},
      eprint={2303.08774},
      archivePrefix={arXiv},
      primaryClass={cs.CL},
      url={https://arxiv.org/abs/2303.08774}, 
}

\end{document}